\def\year{2020}\relax
\documentclass[letterpaper]{article} 
\usepackage{aaai20}  
\usepackage{times}  
\usepackage{helvet} 
\usepackage{courier}  
\usepackage[hyphens]{url}  
\usepackage{graphicx} 
\usepackage{graphicx}  
\usepackage{float, latexsym}
\usepackage{amsfonts,amssymb} 
\usepackage{pgfplots}
\usepackage{tikz}
\usepackage{subfigure}
\usepackage{xcolor}
\usepackage{graphicx}
\usepackage{amsmath}
\usepackage{verbatim}
\title{Video Captioning with Text-based Dynamic Attention and Step-by-Step Learning}
\author{Huanhou Xiao and Jinglun Shi} 
\begin{document}
	
\maketitle

\begin{abstract}
Automatically describing video content with natural language has been attracting much attention in CV and NLP communities. Most existing methods predict one word at a time, and by feeding the last generated word back as input at the next time, while the other generated words are not fully exploited. Furthermore, traditional methods optimize the model using all the training samples in each epoch without considering their learning situations, which leads to a lot of unnecessary training and can not target the difficult samples. To address these issues, we propose a text-based dynamic attention model named TDAM, which imposes a dynamic attention mechanism on all the generated words with the motivation to improve the context semantic information and enhance the overall control of the whole sentence. Moreover, the text-based dynamic attention mechanism and the visual attention mechanism are linked together to focus on the important words. They can benefit from each other during training. Accordingly, the model is trained through two steps: ``starting from scratch" and ``checking for gaps". The former uses all the samples to optimize the model, while the latter only trains for samples with poor control. Experimental results on the popular datasets MSVD and MSR-VTT demonstrate that our non-ensemble model outperforms the state-of-the-art video captioning benchmarks.
\end{abstract}

\begin{figure*}[ht]
	\centering
	\includegraphics[scale=0.2]{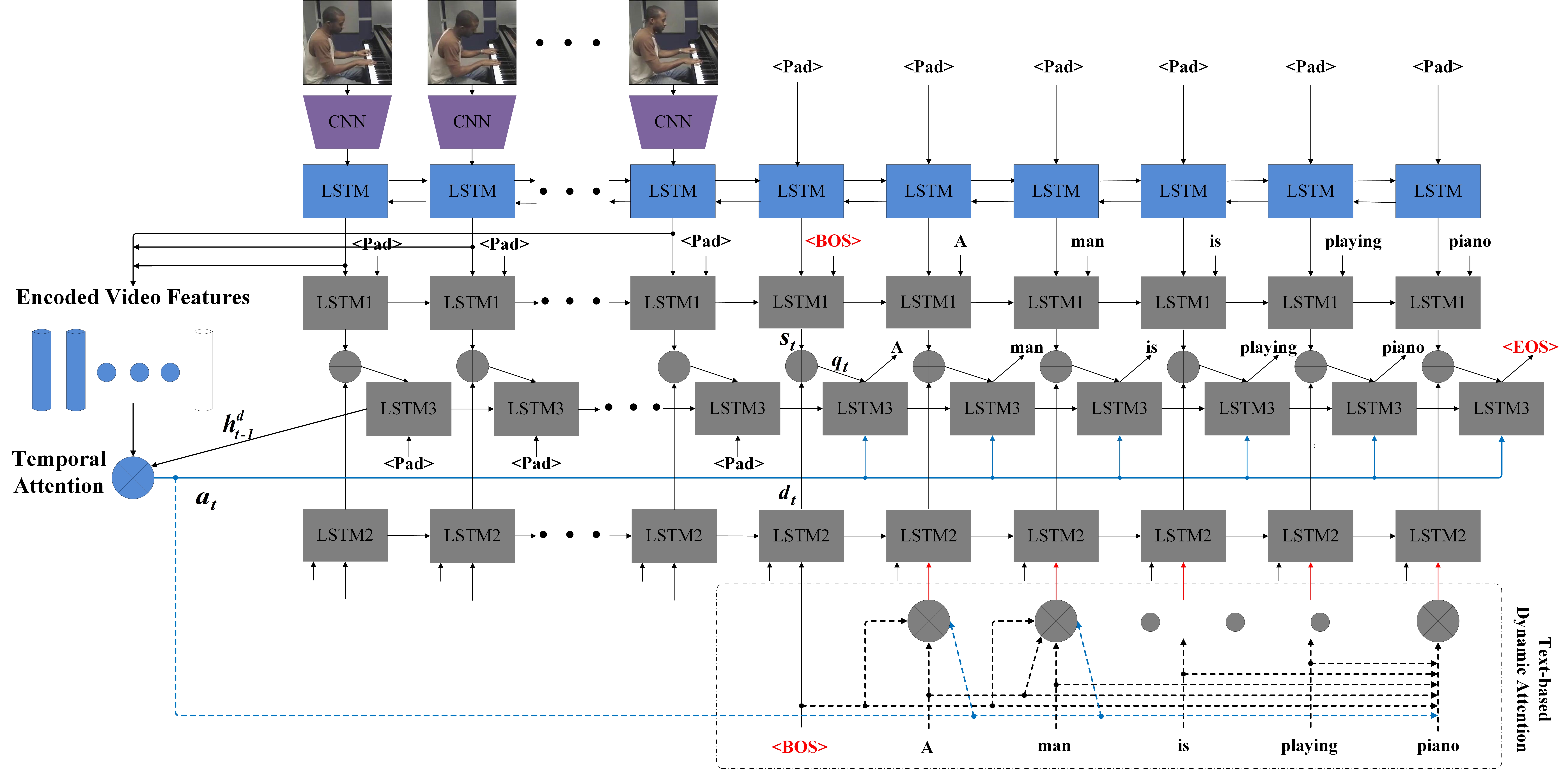}
	\caption{The overall framework of our proposed text-based dynamic attention model. A text-based dynamic attention mechanism is utilized to adaptively focus on all the previously generated words when predicting the next word. Then, the last generated word and the weighted combination of all the previously generated words are used as inputs to two independent LSTMs whose outputs are adjusted by a textual gate to automatically decide their contributions. The proposed TDAM can not only improve the overall control of the whole sentence, but also guide the global visual attention for effective video representation.}
	\label{fig:dynamic_atten_model}
\end{figure*}
\section{Introduction}
Video captioning which translates video into meaningful textual sentences provides the potential to bridge the semantic connection between video and language. A wide range of applications can benefit from it such as human-robot interaction and assist the visually impaired. Early methods~\cite{rohrbach2013translating,lebret2015phrase} primarily generated video description by filling the detected subjects, verbs and objects into a pre-defined template. Inspired by the recent advancements in image classification and machine translation, encoder-decoder framework which combines CNN and RNN to translate the visual input into the textual output has been explored. In this approach, typically a CNN is utilized as an encoder to produce the visual representation and a RNN as a decoder to generate a sequence of words. 

Despite achieving encouraging success in video captioning, previous methods suffer important limitations. Firstly, most LSTM-based models predict the next word by taking as input the last generated word, making other words with the same guiding role underutilized. Although LSTM can update the hidden states adaptively, it is difficult to maintain its superior performance all the time in the transmission of context semantic information, which limits its ability to generate correct description and even leads to some incomplete sentences or repeated generation of a word. Secondly, the distribution of different words is unbalanced in datasets, where distinctive words for describing specific objects are far less than common words such as ``a'' and ``the''. As a result, the common words comprise the majority of training loss (cross entropy loss) and some distinctive words cannot be well learned. Besides, the unbalanced distribution of video categories in datasets leads to over-learning of some categories and ungrasping of others, which results in unsatisfactory performance. Thirdly, previous methods do not take into account the learning situations of the training samples in the training process, leading to many inefficient and useless training.

To address these issues, a text-based dynamic attention model (TDAM) with step-by-step learning is proposed to impose a dynamic attention mechanism on all the words that have been generated during the whole sentence generation process, as shown in Fig. 1. It enables the model to make full use of all the previously generated words and improve the overall control of the whole sentence, which helps our model to predict the next word more accurately. Besides, the visual attention and text-based dynamic attention are linked together to benefit from each other during training. In the mean time, our proposed TDAM is trained progressively, that is, ``starting from scratch" and ``checking for gaps", just like human learning. In the first stage ``starting from scratch", the model does not have any descriptive capabilities, so it is optimized with all the training samples. After the model converges, it will be moved to the second stage ``checking for gaps" whereby samples with poor control are used for training. 

To summarize, the main contributions of this paper are as follows: ($i$) we propose a text-based dynamic attention model with hierarchical LSTM to adaptively utilize all the previously generated words in the next-word generation ($ii$) the visual attention and textual attention are linked to form a closed loop and move forward together. ($iii$)  a step-by-step training method that can effectively optimize the model is introduced. ($iv$) we perform an extensive analysis of our proposed method. Experimental results show that our method achieves superior results compared to existing state-of-the-art models.

\section{Related Work}
As a crucial challenge for visual content understanding, captioning task has attracted much attention for many years. Early works for video captioning mainly focus on rule based systems, which detect the visual elements (subjects, verbs, and objects) firstly, and then generate description using the template-based approach. With the rapid development of deep learning, the encoder-decoder framework has been widely applied to video captioning.~\cite{venugopalan2014translating} transferred knowledge from image caption models via adopting the CNN as the encoder and LSTM as the decoder.~\cite{pan2016jointly} used the mean-pooling caption model with joint visual and sentence embedding. However, they ignore the temporal structures of video. To address this issue,~\cite{yao2015describing} incorporated the local C3D features and a global temporal attention mechanism to select the most relevant temporal segments.~\cite{venugopalan2015sequence} presented a sequence to sequence video captioning model which incorporates a stacked LSTM to read the CNN outputs firstly and then generates a sequence of words. In order to better encode long-range dependencies, a hierarchical recurrent video encoder was designed by~\cite{pan2016hierarchical} to exploit multiple time-scale abstraction of the temporal information.~\cite{baraldi2017hierarchical} proposed an LSTM cell that identifies discontinuity points between frames or segments and modifies the temporal connections of the encoding layer accordingly. 

More recently, to generate a high-quality description for a target video, the authors in~\cite{gao2017video} presented a multimodal embedding approach to map the video features and sentence vectors into a joint space to guarantee the semantic consistency of the sentence description and video visual content.~\cite{song2017hierarchical} designed an adjusted temporal attention mechanism to avoid focusing visual attention on non-visual words during caption generation.~\cite{Aafaq_2019_CVPR} embedded temporal dynamics in visual features by hierarchically applying Short Fourier Transform to CNN features. In~\cite{wang2018reconstruction}, a novel encoder-decoder-reconstruction network was proposed to utilize both the forward and backward flows for video captioning. 

Although the above video captioning approaches achieve excellent results, limitations still exist. During training, these models do not take into account the learning situations of the training samples. Besides, a potential disadvantage of them is that the previously generated words are not fully exploited when predicting the next word. Thus, in this paper, a text-based dynamic attention model with step-by-step learning is proposed to focus on all the previously generated words. Simultaneously, the step-by-step training method enables our model to target the undereducated parts and have higher learning efficiency. To alleviate the problems caused by unbalanced distribution of words and video categories, a sentence-level loss consisting of evaluation metrics is designed in the second step. We combine the cross-entropy loss and reward-based loss (sentence-level loss) as a mixed-loss during ``checking for gaps" stage, which takes advantages of two training methods and achieves an improved balance.
\section{Approach}
In this section, we present our approach for video captioning. The baseline temporal attention model is first demonstrated. Then, we propose our text-based dynamic attention model. Finally, our step-by-step learning method which divides the training process into two steps and simultaneously considers word-level loss and sentence-level loss is introduced. In addition, solution details are provided.
\subsection{Baseline Temporal Attention Model}

Our baseline temporal attention model is similar to the standard machine translation encoder-decoder LSTM model. Given a video, \textbf{V}, with \emph{N} frames, the extracted visual features and the embedded textual features can be represented as \textbf{v}={\{$\emph{v}_{1}$,$\emph{v}_{2}$,...,$\emph{v}_{N}$\}} and \textbf{w}={\{$\emph{w}_{1}$,$\emph{w}_{2}$,...,$\emph{w}_{T}$\}}, where $\emph{v}_{i}$ $\in$ $\mathbb{R}^{D_{v}\times 1}$, $\emph{w}_{i}$ $\in$ $\mathbb{R}^{D_{w}\times 1}$, and \emph{T} is the sentence length. Specifically, ${D}_{v}$ and ${D}_{w}$ are the respective dimensions of the frame-level features and vocabulary. We use a bi-directional LSTM (Bi-LSTM)~\cite{graves2005framewise} which can capture both forward and backward temporal relationships to encode the extracted visual features, and a visual attention mechanism is incorporated following it. To avoid imposing visual attention on non-visual words~\cite{song2017hierarchical}, we append a blank feature whose values are all zeros to the encoded video features. Thus, we can eliminate the impact of visual information if the predicting word is irrelevant to the high-level visual representations. Accordingly, the output context vector at time step \emph{t} can be represented as:
\begin{small}
	\begin{equation}
	a_t=\sum_{i=1}^{N+1}\alpha_{t,i}h_i
	\end{equation}
\end{small}
where $h_{i,\,i\in [1,N]}$ is the hidden state of the Bi-LSTM, $h_{N+1}$ is the blank feature, and $\alpha_{t,i}$ is the attention weight which can be computed as:
\begin{equation}
	\alpha_{t,i}={\rm softmax}(e_{t,i})
\end{equation}
\begin{equation}
	e_{t,i}=w^T{\rm tanh}(W_ah_i+V_ah^d_{t-1}+b_a)
\end{equation}
where $w$, $W_a$, $V_a$ and $b_a$ are the learned parameters, $h^d_{t-1}$ is the hidden state of the decoder LSTM at the (\emph{t}-1)-th time step. Formally, the distribution of the output sequence with respect to the input sequence is:
\begin{equation}
	p(w_1,...,w_T|v_1,...,v_N)=\prod_{i=1}^{T}p(w_t|w_{<t},\textbf{V};\theta)
\end{equation}
where $\theta$ is the model parameter set, and the distribution $p(w_t|w_{<t},\textbf{V};\theta)$ is given by softmax over all the words in the vocabulary.

\subsection{Text-based Dynamic Attention Model}
Based on the temporal attention model introduced above, we propose in this subsection a text-based dynamic attention model for video captioning. Using a dynamic attention mechanism, we can make full use of all the previously generated words when predicting the next word, as shown in Fig. 1. At time step $t$, suppose there are $(t-1)$ words have been generated, and the output of the visual attention and the text-based dynamic attention are $a_t$ and $\bar{w}_t$ respectively. Then $\bar{w}_t$ can be represented as:
\begin{equation}
	\bar{w}_t=\sum_{i=0}^{t-1}\beta_{t,i}w_i
\end{equation}
\begin{equation}
	\beta_{t,i}={\rm softmax}(u_{t,i})
\end{equation}
\begin{equation}
	u_{t,i}=\bar{w}^T{\rm tanh}(\bar{W}w_i+\bar{V}a_t+\bar{b})
\end{equation}
where $\bar{w}$, $\bar{W}$, $\bar{V}$ and $\bar{b}$ are the learned parameters, and $w_0$ represents the embedded vector of the begin-of-sentence $<$BOS$>$ tag. Considering that LSTM can transmit context information in most cases, we use LSTM1 which encodes the last generated word as the dominant component and LSTM2 which encodes the combination of all the previously generated words as the auxiliary to capture textual information together. Note that the other inputs of LSTM2 are the same as LSTM1. Specifically, we design an adjusted gate to automatically adjust the outputs from these two LSTMs. The adjusted gate is calculated as follow:
\begin{equation}
	g_t={\rm sigmoid}(W_sh^{LSTM\emph{1}}_t)
\end{equation}
where $W_s$ is learned parameter, and $h^{LSTM\emph{1}}_t$ denotes the hidden state of the LSTM1. Obviously, $g_t$ is projected onto the range of [0; 1]. Suppose the input context vector of the LSTM3 is $q_t$, and the outputs of the LSTM1 and LSTM2 are $s_t$ and $d_t$, respectively. Then, $q_t$ can be calculated as:
\begin{equation}
	q_t=g_ts_t+(1-g_t)d_t
\end{equation}

In our presented TDAM, the visual attention and text-based attention are linked together by using the weighted visual feature to compute the contributions of all the generated words, which helps to concentrate on some important words and improve the prediction of the next word. As a result, the correct predictive word can better guide the visual attention in turn at the next time. Repeatedly, two attention mechanisms can benefit from each other and move forward together, which improves the video description.
\begin{figure}[t]
	\centering
	\includegraphics[scale=0.2]{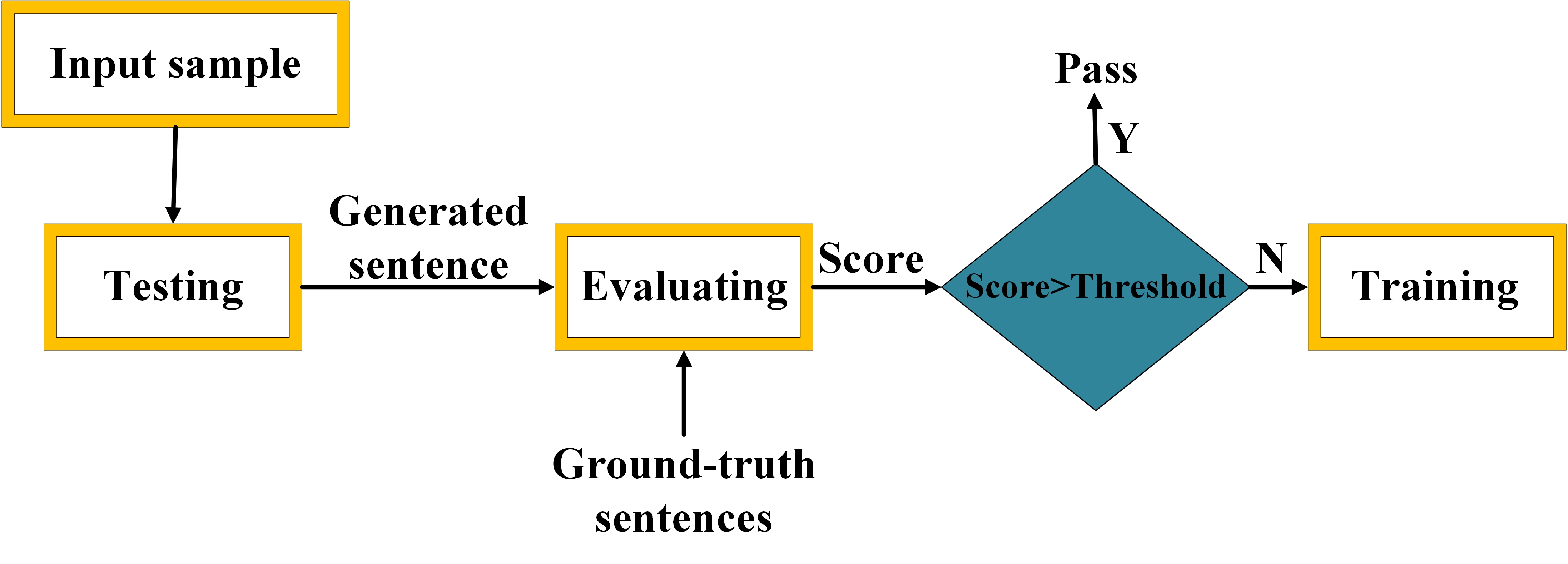}
	\caption{The flow diagram of our step2. For each training sample, we first test it for evaluation. If the evaluation score is higher than the threshold we set, no training is required, otherwise the input sample will be used to optimize the model.}
	\label{step-by-step}
\end{figure}
\subsection{Step-by-Step Learning}
As described in the previous section, we progressively train our model in two steps. The first step aims to have a warm start so that the model is optimized with all the training samples. The second step mainly focuses on the poorly controlled samples, thus the model can make up for these shortcomings.

\textbf{Step1: “starting from scratch”}. In the first step, our model, like a newborn child, does not have any descriptive ability. Therefore, all the training samples are used to optimize the model. Considering that the high resemblance between the generated sentences and the training sentences helps the model to converge faster, step1 is the standard training method using only the cross entropy loss. As introduced above, the loss function of the network can be defined as:
\begin{equation}
	L_{xe}=-\sum_{t=1}^{T}{\rm log}P(w_t|w_{<t},\textbf{V};\theta)
\end{equation}

\textbf{Step2: “checking for gaps”}. After step1, the model has been able to correctly describe most of the video contents that belong to training samples. It is more effective at this stage to target what has not been mastered (e.g. a small number of descriptive words in training samples), rather than learning all samples repeatedly. Thus, before deciding whether to train the input samples, we need to evaluate them, as described in Fig. 2. In our experiments, an evaluated score is designed to assess whether the training sample is mastered. It can be computed as follows:
\begin{equation}
\rm {Score=BLEU-4+ROUGE}
\end{equation}

Here, BLEU-4~\cite{papineni2002bleu} and ROUGE~\cite{lin2004rouge} are the sentence evaluation metrics, which can well assess the quality of generated sentence. BLEU-n matches words and computes n-gram precision between the reference set and candidate sentence. It captures the lexical and textual consistency between sentences. Therefore, a high BLEU-4 score largely means that the model can generate descriptive words correctly. As for ROUGE, it considers the recall which complements BLEU. Another important reason we choose them as evaluated score and sentence reward (this will be introduced later) is that the time cost is much less than using other metrics (e.g. METEOR~\cite{denkowski2014meteor}). If the evaluated score of the input sample is higher than the threshold we set, we believe that this sample has been well mastered by the model and will not be retrained. Otherwise, training will continue. Besides, in this step, instead of only utilizing the word-level cross entropy loss which encourages high resemblance to the corresponding ground-truth caption and suppresses other reasonable descriptions, a reward-based reinforcement learning loss is designed to encourage the diversity of the generated descriptions and focus more on the descriptive words. To directly optimize the reward-based loss, a policy gradient method is utilized. Concretely, our text-based dynamic attention model acts as an agent and interacts with an external environment (video and sentence). We define a policy, $p_{\theta}$, which influences the action that generates the next word. After the whole sentence is generated, we can compute a reward based on it and the ground-truth sentences. The reward is defined as the combination of evaluated metrics, the same as the evaluated score introduced above. Our training objective is to minimize the negative expected reward:

\begin{equation}
	L_{re}(\theta)=-\mathbb{E}_{w^s\sim p_{\theta}}[r(w^s)]
\end{equation}
where $w^s$ is the word sequence sampled from the model. Based on the REINFORCE algorithm, the expected gradient of a non-differentiable reward function can be computed as follows:
\begin{equation}
	\nabla_{\theta}L_{re}(\theta)=-\mathbb{E}_{w^s\sim p_{\theta}}[r(w^s)\nabla_{\theta}{\rm log}p_{\theta}(w^s)]
\end{equation}
the above gradient can be approximated using a single Monte Carlo sample $w^s$ from $p_\theta$ as follows:
\begin{equation}
	\nabla_{\theta}L_{re}(\theta)=-r(w^s)\nabla_{\theta}{\rm log}p_{\theta}(w^s)
\end{equation}

However, the above approximation has a high variance because of the gradient estimate. To reduce this variance, we baseline the REINFORCE algorithm with the reward obtained by the current model under the reference algorithm used at test time~\cite{rennie2017self}. Here we choose the greedy search of words as our baseline. Suppose the reward obtained by the current model under greedy search is $r(w^b)$. Then, the expected gradient can be rewritten as:
\begin{equation}
	\nabla_{\theta}L_{re}(\theta)=-(r(w^s)-r(w^b))\nabla_{\theta}{\rm log}p_{
		\theta}(w^s)
\end{equation}
using the chain rule, we have:
\begin{equation}
	\nabla_{\theta}L_{re}(\theta)=\sum_{t=1}^{T}\frac{\partial L_{re}(\theta)}{\partial s_t}\frac{\partial s_t}{\partial \theta}
\end{equation}
where $s_t$ is the input to the softmax function, and $\frac{\partial L_{re}(\theta)}{\partial s_t}$ is given by ~\cite{zaremba2015reinforcement}:
\begin{equation}
	\frac{\partial L_{re}(\theta)}{\partial s_t}\approx(r(w^s)-r(w^b))(p_{\theta}(w_t|h^d_t-1_{w^s_t}))
\end{equation}

Here, $h^d_t$ is the hidden state of the decoder. Accordingly, the words sampled from the model that return a higher reward than $w^b$ will be “encouraged” by increasing their word probabilities, while the words that return a lower reward will be “discouraged” by decreasing their word probabilities.

To maintain the advantages of both losses and ensure the readability of the generated caption, in step2 we use a mixed loss function as our training objective function, which is a weighted combination of the cross entropy loss and reinforcement learning loss:
\begin{equation}
	L=\lambda L_{xe}+(1-\lambda)L_{re}
\end{equation}
where $\lambda$ represents a tuning parameter that is used to balance them.
\section{Experiments}
\label{sec:experiments}

\subsection{Datasets}
We evaluate our model on the widely used Microsoft Video Description (MSVD) corpus~\cite{chen2011collecting} and the MSR-Video to Text (MSR-VTT) dataset~\cite{xu2016msr}. The MSVD dataset consists of 1,970 video clips collected from YouTube, which covers a lot of topics and is well-suited for training and evaluating a video captioning model. We adopt the same data splits as provided in~\cite{venugopalan2015sequence} with 1,200 videos for training, 100 videos for validation and 670 videos for testing. Regarding the MSR-VTT dataset, there are 10K video clips and 20 reference sentences annotated by human are provided for each video clip. We follow the public split method: 6,513 videos for training, 497 videos for validation, and 2,990 videos for testing.
\subsection{Experimental Settings}
We uniformly sample 60 frames from each clip and use Inception-v3~\cite{szegedy2016rethinking} to extract frame-level features. To capture the temporal information and audio information of video, the pre-trained C3D network~\cite{karpathy2014large} and VGGish model~\cite{hershey2017cnn} are utilized to extract the dynamic features of video and process the raw WAV files extracted from video, respectively. We convert all the sentences to lower cases, remove punctuation characters and tokenize the sentences. We retain all the words in the dataset and thus obtain a vocabulary of 13,375 words for MSVD, 29,040 words for MSR-VTT. To evaluate the performance of our model, we utilize METEOR~\cite{denkowski2014meteor}, BLEU~\cite{papineni2002bleu}, CIDEr~\cite{vedantam2015cider} and ROUGE~\cite{lin2004rouge} as our evaluation metrics, which are commonly used for performance evaluation of video captioning methods.

In our experiments, we add a begin-of-sentence $<$BOS$>$ tag at the beginning of the caption, and an end-of-sentence tag $<$EOS$>$ to its end, so that our model can handle captions with varying lengths. In addition, with an initial learning rate $10^{-5}$ to avoid the gradient explosion, the LSTM unit size and word embedding size are set as 512, empirically. Our objective function is optimized using the ADAM optimizer~\cite{kingma2014adam}. We train our model with mini-batch 64, and the length of sentence \emph{T} is set as 20. For sentence with fewer than 20 words, we pad the remaining inputs with zeros. To regularize the training and avoid overfitting, we apply dropout with rate of 0.5 on the outputs of LSTMs. For step-by-step learning, the evaluated threshold is set as 1.9. During testing process, beam search with beam width of 5 is used to generate descriptions.
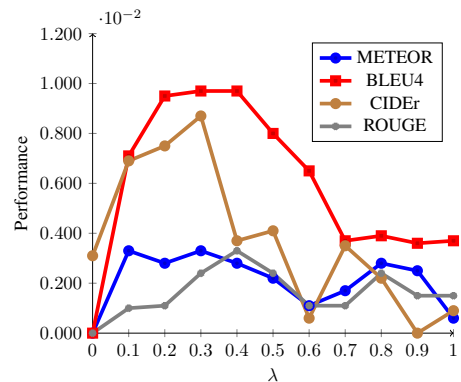
\begin{figure}[t]
	\centering
	\begin{tikzpicture}[scale=0.7]
	\begin{axis}[
	axis lines = left, 
	axis line style={->},
	ylabel near ticks,
	xlabel near ticks,
	ymin=-0, ymax=0.012,
	xtick={0,0.1,0.2,0.3,0.4,0.5,0.6,0.7,0.8,0.9,1.0},
	ytick={0,0.002,0.004,0.006,0.008,0.01,0.012},
	y tick label style={
		/pgf/number format/.cd,
		fixed,
		fixed zerofill,
		precision=3,
		/tikz/.cd
	},
	legend pos=north east,
	legend style={legend columns=1},
	xlabel = $\lambda$,
	ylabel = {Performance}]
	\addplot+[draw=blue,line width=2] coordinates {(0,0) (0.1,0.0033) (0.2,0.0028) (0.3,0.0033) (0.4,0.0028) (0.5,0.0022) (0.6,0.0011) (0.7,0.0017) (0.8,0.0028) (0.9,0.0025) (1.0,0.0006)};
	\addlegendentry{METEOR}
	\addplot+[draw=red,line width=2] coordinates {(0,0)  (0.1,0.0071) (0.2,0.0095) (0.3,0.0097) (0.4,0.0097) (0.5,0.0080) (0.6,0.0065) (0.7,0.0037) (0.8,0.0039) (0.9,0.0036) (1.0,0.0037)};
	\addlegendentry{BLEU4}
	\addplot+[draw=brown,line width=2] coordinates {(0,0.0031)  (0.1,0.0069) (0.2,0.0075) (0.3,0.0087) (0.4,0.0037) (0.5,0.0041) (0.6,0.0006) (0.7,0.0035) (0.8,0.0022) (0.9,0) (1.0,0.0009)};
	\addlegendentry{CIDEr}
	\addplot+[draw=gray,line width=2] coordinates {(0,0)  (0.1,0.001) (0.2,0.0011) (0.3,0.0024) (0.4,0.0033) (0.5,0.0024) (0.6,0.0011) (0.7,0.0011) (0.8,0.0024) (0.9,0.0015) (1.0,0.0015)};
	\addlegendentry{ROUGE}
	\end{axis}
	\end{tikzpicture}
	\caption{Effect of $\lambda$}
	\label{fig:effect}
\end{figure}
\begin{table*}[t]
	\begin{center}
		\caption{Captioning performance comparison on MSVD and MSR-VTT datasets. R, I, Iv4 and IRv2 denotes ResNet, Inception-v3, Inception-v4 and Inception-ResNet-v2, respectively. M denotes multiple features, NL denotes two attentions are not linking, and S denotes the model is trained step-by-step. B-4 denotes BLEU4 metric. Note that audio is not available on MSVD dataset. The symbol ``-" indicates such metric is unreported.} \label{tab:Captioning performance comparison on MSVD and MSR-VTT}
		\begin{tabular}{c|c c c c|c c c c}
			\hline
			&\multicolumn{4}{c}{MSVD}&\multicolumn{4}{c}{MSR-VTT}\\
			\hline
			Model &METEOR & B-4 & CIDEr & ROUGE &METEOR & B-4 & CIDEr & ROUGE \\
			\hline
			\hline
			S2VT~\cite{venugopalan2015sequence} (M) & 29.8 & - & - & - & - & - & - & -\\
			HRNE~\cite{pan2016hierarchical} (M) & 33.9 & 46.7 & - & - & - & - & - & -\\
			MS-RNN~\cite{song2018deterministic} (R) & 33.8 & 53.3 & 74.8 & 70.2 & 26.1 & 39.8 & 40.9 & 59.3\\
			aLSTMs~\cite{gao2017video} (I) & 33.3 & 50.8 & 74.8 & - & 26.1 & 38.0 & 43.2 & -\\
			hLSTMat~\cite{song2017hierarchical} (R) & 33.6 & 53.0 & 73.8 & - & 26.3 & 38.3 & - & -\\
			RecNet~\cite{wang2018reconstruction} (Iv4) & 34.1 & 52.3 & 80.3 & 69.8 & 26.6 & 39.1 & 42.7 & 59.3\\
			E2E~\cite{li2019end} (IRv2) & 34.1 & 50.3 & 87.5 & 70.8 & 27.0 & 40.4 & 48.3 & 61.0\\
			GRU-EVE~\cite{Aafaq_2019_CVPR} (M) &35.0 & 47.9 & 78.1 & 71.5 & 28.4 & 38.3 & 48.1 & 60.7\\
			MARN~\cite{pei2019memory} (M) &35.1 & 48.6 & \textbf{92.2} & 71.9 & 28.1 & 40.4 & 47.1 & 60.7\\
			v2t-navigator~\cite{jin2016describing} (M) & - & - & - & - & 28.2 & 40.8 & 44.8 & 60.9\\
			VideoLAB~\cite{ramanishka2016multimodal} (M) & - & - & - & - & 27.7 & 39.1 & 44.1 & 60.6\\
			Aalto~\cite{shetty2016frame} (M) & - & - & - & - & 26.9 & 39.8 & 45.7 & 59.8\\
			\hline
			\hline
			Baseline Attention Model (I) & 35.0 & 51.4 & 82.1 & 70.8 & 26.4 & 37.7 & 42.3 & 58.4\\
			Baseline deep LSTM Model (I) & 34.5 & 51.0 & 79.7 & 70.7 & 26.9 & 38.5 & 43.2 & 59.0\\
			TDAM (I) & 35.2 & 52.6 & 85.0 & 71.2 & 27.3 & 39.4 & 44.9 & 59.7\\
			TDAM (I-NL) & 35.3 & 52.4 & 84.7 & 71.6 & 27.1 & 38.5 & 42.9 & 58.8\\
			TDAM (M) & 35.8 & 53.6 & 85.1 & 72.1 & \textbf{28.7} & 43.5 & 48.3 & 61.9\\
			TDAM (M-NL) & 35.4 & 53.0 & 82.8 & 71.8 & 28.5 & 41.9 & 46.4 & 61.0\\
			TDAM (M-S) & \textbf{36.1} & \textbf{54.0} & 85.8 & \textbf{72.3} & \textbf{28.7} & \textbf{44.7} & \textbf{48.9} & \textbf{62.3}\\
			\hline
		\end{tabular}
	\end{center}
\end{table*}
\subsection{Ablation studies}
\subsubsection{Effect of Text-based Dynamic Attention}
To focus on all the previously generated words in the next-word generation, a text-based dynamic attention method is utilized. From the second block of Table 1, we observe that our proposed TDAM (I) outperforms our strong baseline attention model, showing the effectiveness of our text-based dynamic attention. To further verify whether the performance enhancement is caused by our ingenious design or just the stacking of multiple LSTMs, based on the attention model, we train an additional model with the same LSTM layers as TDAM. We refer to it as the baseline deep LSTM model. We notice that the baseline deep LSTM model behaves differently on two datasets, which may be related to the different distribution of video categories in datasets. Nevertheless, our model consistently surpasses it on both datasets. This indicates the superiority of our TDAM.
\subsubsection{Effect of Linking Two Attention Mechanism}
In our TDAM, the visual temporal attention mechanism and the text-based dynamic attention mechanism are linked together to focus on some important vectors and improve the prediction of the next word. In addition, two attention mechanisms can benefit from each other and move forward together, as described earlier. The comparison results of linking and unlinking two attentions are reported in Table 1, From which we can observe that the model with linked attentions shows superior performance, especially when using multiple features.
\subsubsection{Effect of $\lambda$}
To achieve a better balance between the cross-entropy loss and the sentence-level loss, we study the performance variance with different $\lambda$ in these sub-experiments. We tune $\lambda$ from 0 to 1 at intervals of 0.1 on the MSVD dataset, and the results are reported in Fig. 3. At this point, we normalize our evaluation scores using the following function: 
\begin{equation}
	Q_{norm}=\frac{Q-min(Q)}{min(Q)}
\end{equation}
where $Q$ and $Q_{norm}$ are the original and normalized performance values, respectively.

Fig. 3 shows that, when $\lambda=0.3$, our model maintains a better balance between the cross-entropy loss and the sentence-level loss. Besides, the effect of mixed loss is obviously better than the single loss. Thus, in the following experiments, we set $\lambda=0.3$.

\subsection{Comparison with the State-of-the-Art}
\subsubsection{Quantitative Analysis}
Table 1 demonstrates the result comparison among our proposed method and some state-of-the-art models. The comparing algorithms include the encoder-decoder based architectures (S2VT~\cite{venugopalan2015sequence}, MS-RNN~\cite{song2018deterministic}, E2E~\cite{li2019end}), and the attention based methods (HRNE~\cite{pan2016hierarchical}, aLSTMs~\cite{gao2017video}, hLSTMat~\cite{song2017hierarchical}, RecNet~\cite{wang2018reconstruction}, GRU-EVE~\cite{Aafaq_2019_CVPR}, MARN~\cite{pei2019memory}). For MSVD dataset, as can be seen from Table 1, compared with the models using single feature or multiple features, both TDAM (I) and TDAM (M-S) perform best on most metrics, verifying the superiority of our proposed approach. Specifically, TDAM (M-S) surpasses the best counterpart (i.e., MARN) by 1.0$\%$, 5.4$\%$, and 0.4$\%$ for METEOR, BLEU-4, and ROUGE respectively. In addition, we notice that TDAM (M-S) performs better than TDAM (M). This indicates that it is beneficial to train our model using step-by-step learning.
\begin{figure*}[t]
	\centering
	\includegraphics[scale=0.25]{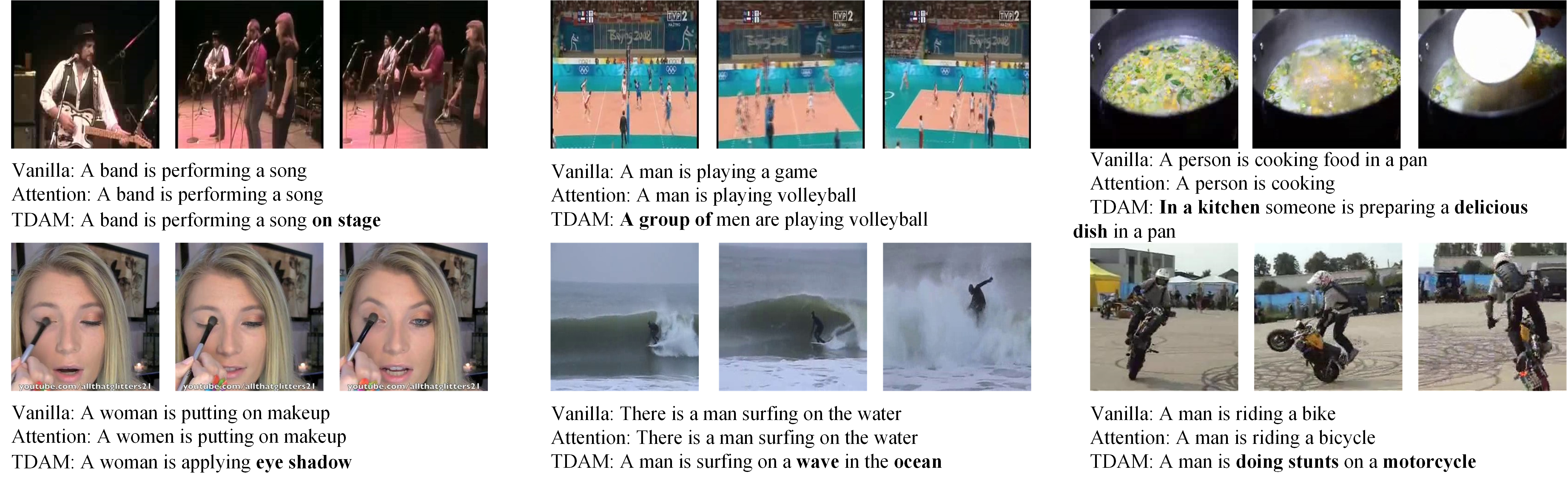}
	\caption{Examples of video captioning.}
	\label{fig:Examples of video captioning}
\end{figure*}

For MSR-VTT, we also compare our models with the top-3 results from the MSR-VTT challenge in the table$\footnote{http://ms-multimedia-challenge.com/.}$, including v2t-navigator~\cite{jin2016describing}, Aalto~\cite{shetty2016frame} and VideoLAB~\cite{ramanishka2016multimodal}, which are all based on features from multiple cues such as action features and audio features. The experimental results presented in Table 1 show that our TDAM performs significantly better than other methods on all metrics. We observe that in terms of CIDEr metric, our model performs better than MARN on MSR-VTT, but achieves worse performance on MSVD. We think this is related to the distribution of the video categories and the annotation sentences in the database. Because CIDEr regards each sentence as a ``document" and represents it as a form of TF-IDF vector, and then calculates the cosine similarity between the reference captions and the candidate captions. But in general, our TDAM is superior to MARN.

\subsubsection{Qualitative Analysis}
To gain an intuition of the improvement on generated video descriptions of our proposed TDAM, we present some video examples with the video description from vanilla model and temporal attention model as comparison to our model in Fig. 4. From the upper row of Fig. 4, we can observe that our TDAM generates more accurate and detailed descriptions for diverse video topics than the vanilla and attention model. In the bottom row of Fig. 4, even though all the caption models generate correct descriptions, the generated sentences from our model are more discriminative to describe the video contents.

Fig. 5 shows a few example sentences along with the values of adjusted gate. The value reflects the effects of LSTM1 which takes input the last generated word and LSTM2 which takes input the combination of all the previously generated words when predicting each word. The larger the value, the greater the effect of LSTM1, and vice versa. It can be seen that the information contained in LSTM1 is generally sufficient to generate the next word, especially in the case of generating short sentences (e.g. the first video), which proves the ability of LSTM to transmit context information. However, in some cases, the model needs to explicitly utilize other generated words. For example, in the last video, we observe that LSTM2 plays a major role when generating the word ``stadium". This is because the word ``stadium" needs to be inferred from ``two men are playing table tennis". It is predicted by relying on other previously generated words. This is also the case with the second video. The model infers ``kitchen" from ``a person is cutting a piece of meat", which indicates that making full use of all the previously generated words is essential.
\begin{figure}[t]
	\centering
	\includegraphics[width=\columnwidth]{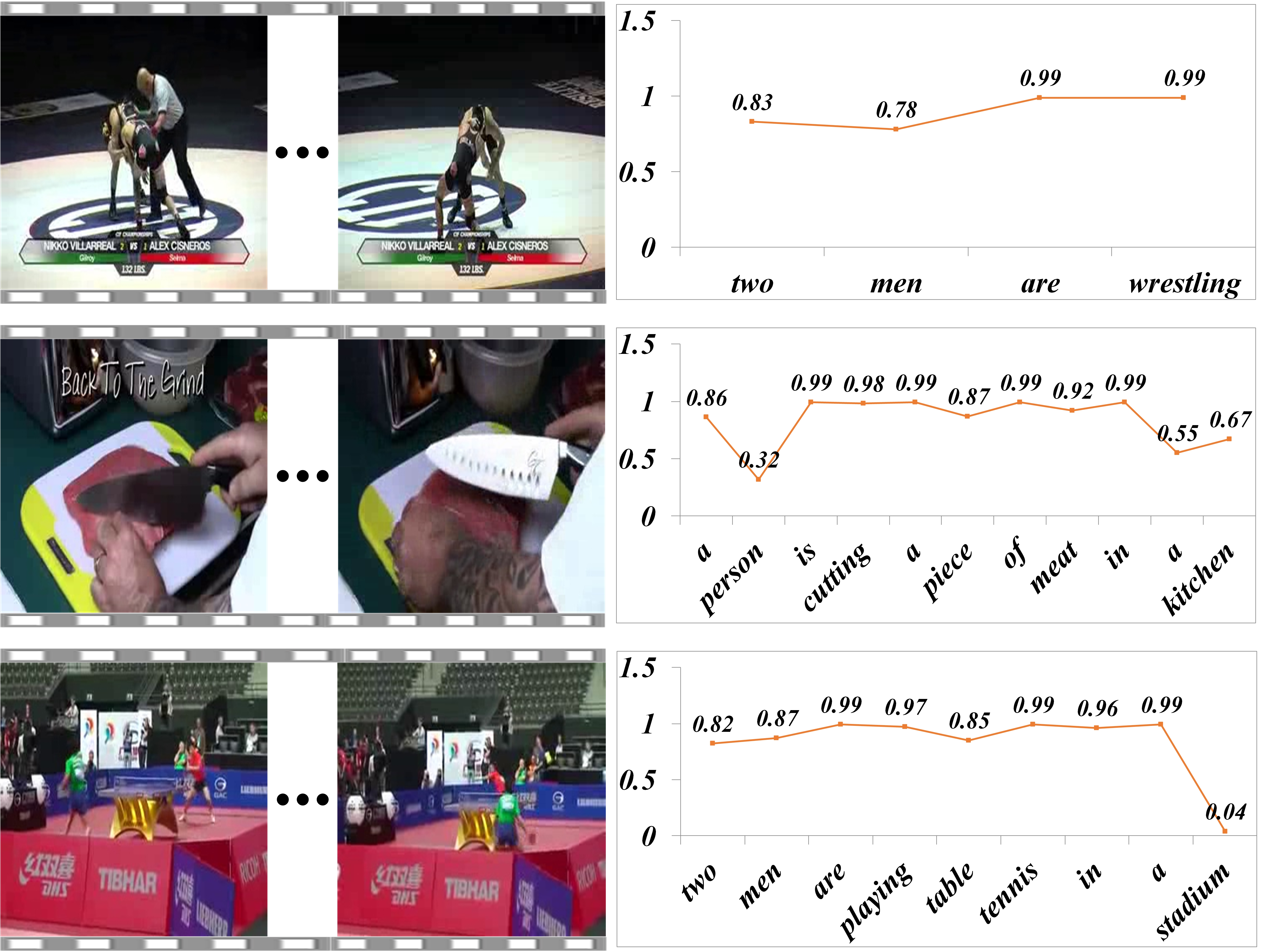}
	\caption{Visualization of the values of adjusted gate aligned with input video and generated caption.}
	\label{fig:Visualization of generated captions}
\end{figure}
\section{Conclusion}
In this paper, we propose a text-based dynamic attention model with step-by-step learning, which flexibly makes use of all the previously generated words in the caption generation process. Furthermore, the visual temporal attention mechanism and the text-based dynamic attention mechanism are linked together by using the weighted visual feature to guide the textual attention, which is conducive to concentrate on the important words when predicting the next word, and benefits the visual attention in turn. In the step2, our model is implemented by simultaneously minimizing the word-level cross entropy loss and sentence-level reinforcement learning loss. With all the designs, we achieve the superior results on both MSVD and MSR-VTT. Our future work will consider designing a better reward for our step-by-step learning.


\bibliographystyle{aaai}
\bibliography{refs}

\end{document}